\documentclass{article}
\usepackage{spconf,amsmath,graphicx}
\usepackage{booktabs}
\usepackage{xcolor}

\title{Approaches Toward Physical and General Video Anomaly Detection}
\name{Laura Kart and Niv Cohen}
\address{School of Computer Science and Engineering\\
The Hebrew University of Jerusalem, Israel.\\}
\begin{document}
\ninept
\maketitle
\begin{abstract}
In recent years, many works have addressed the problem of finding never-seen-before anomalies in videos. Yet, most work has been focused on detecting anomalous frames in surveillance videos taken from security cameras. Meanwhile, the task of anomaly detection (AD) in videos exhibiting anomalous mechanical behavior, has been mostly overlooked. Anomaly detection in such videos is both of academic and practical interest, as they may enable automatic detection of malfunctions in many manufacturing, maintenance, and real-life settings. To assess the potential of the different approaches to detect such anomalies, we evaluate two simple baseline approaches: (i) Temporal-pooled image AD techniques. (ii) Density estimation of videos represented with features pretrained for video-classification.

Development of such methods calls for new benchmarks to allow evaluation of different possible approaches. We introduce the Physical Anomalous Trajectory or Motion (PHANTOM) dataset\footnote{https://github.com/laurarkart/Physical-Anomalous-Trajectory-or-Motion-PHANTOM-Dataset}, which contains six different video classes. Each class consists of normal and anomalous videos. The classes differ in the presented phenomena, the normal class variability, and the kind of anomalies in the videos. We also suggest an even harder benchmark where anomalous activities should be spotted on highly variable scenes. 
\end{abstract}

\begin{keywords}
Anomaly Detection, Video Anomaly Detection
\end{keywords}

\section{Introduction}

Detecting never-seen-before novelties, or anomalies, is a key ability humans use to raise awareness of new dangers and opportunities. Examples of such include spotting new behaviors in natural systems, detecting security threats, or spotting equipment malfunctions. 
While anomaly detection (AD) methods aimed at images \cite{reiss2021panda,hendrycks2019using,tack2020csi} usually address a large variety of data domains, video AD has been mostly focused on surveillance videos. In such videos \cite{luo2017revisit,mahadevan2010anomaly}, typically taken from security cameras, normal walking behavior is usually defined as the normal class, and other behaviors (riding a bike, crowd gathering) are defined as anomalies. In recent years, many new techniques have been suggested, pushing forward performance on such benchmarks \cite{georgescu2021anomaly,park2020learning}. Nevertheless, other kinds of video anomalies remain mostly out of the scope of contemporary research. 


In this work we explore different kinds of anomalies, examining different types of methods to address them. Our work addresses physical abnormalities in the motion or trajectory of an object that have been largely overlooked and presents a new benchmarks for video anomaly detection.

\begin{figure}[t]
\begin{center}
\begin{tabular}{@{\hskip0pt}c@{\hskip5pt}c@{\hskip5pt}c@{\hskip5pt}c@{\hskip5pt}c@{\hskip5pt}c@{\hskip0pt}c@{\hskip0pt}c@{\hskip0pt}c}

Candle & Keyboard & Sushi  \\

\includegraphics[width=0.3\linewidth]{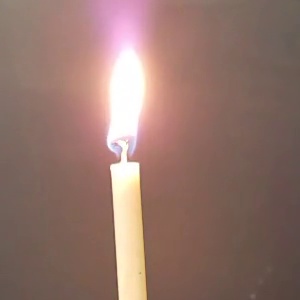} &
\includegraphics[width=0.3\linewidth]{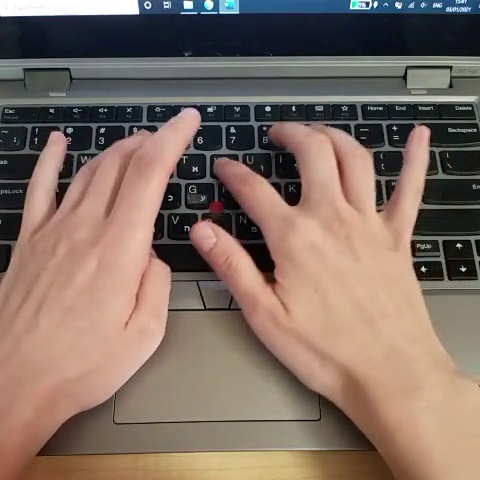} &
\includegraphics[width=0.3\linewidth]{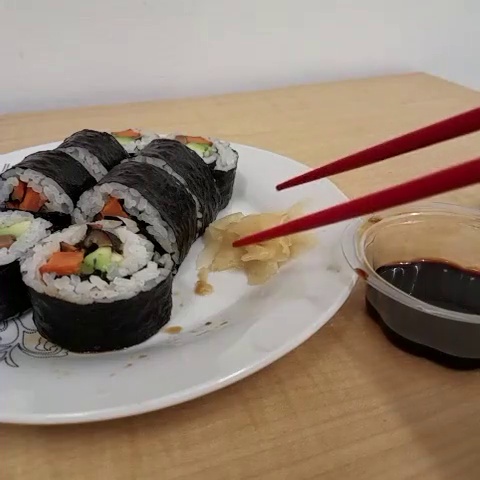} &
\\
Magnets & Pendulum & Window \\

\includegraphics[width=0.3\linewidth]{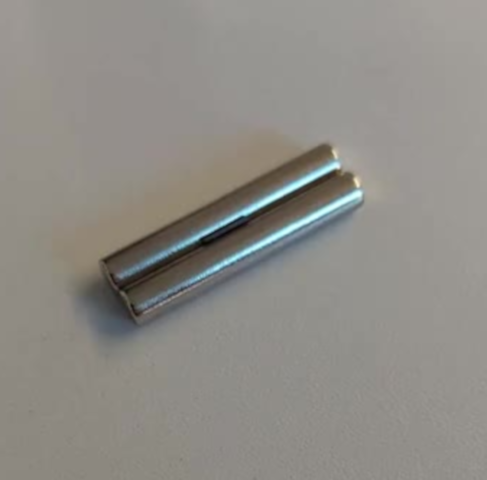} &
\includegraphics[width=0.3\linewidth]{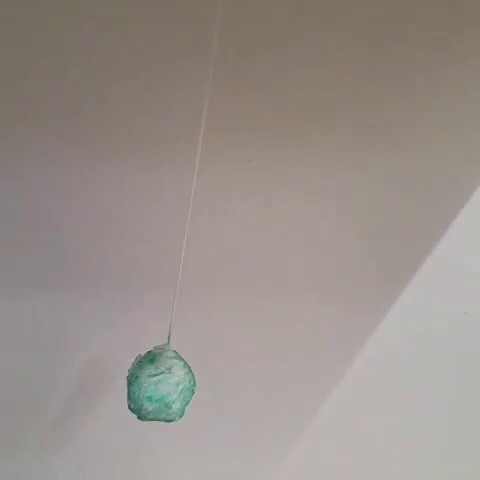} &
\includegraphics[width=0.3\linewidth]{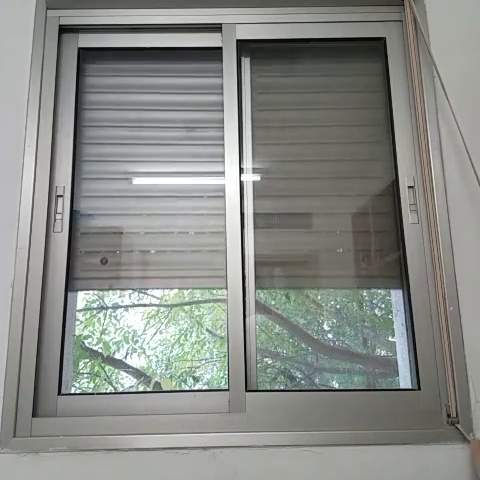} &
\\

\end{tabular}
\end{center}
\caption{Examples of frames from each of the six video types in the PHANTOM dataset.}
\label{fig:ffhq}
\vspace{-1em}
\end{figure}

\section{Related Work}
\label{sec:related_works}
\textbf{Video Anomaly Detection}
As most work on video anomaly detection has focused on surveillance videos, various methods have been tried to tackle this problem. Deep learning methods have been able to make significant progress toward solving this problem, using auto-encoders \cite{ionescu2019object,gong2019memorizing}, generative models \cite{hasan2016learning}, or prediction \cite{lee2019bman}. More recent methods have been able to outperform using pretraining and self-supervised learning \cite{pang2020self,georgescu2021anomaly,park2020learning}.

\textbf{Image Anomaly Detection}
Image anomaly detection is a fast-growing field, often using similar techniques to those used in video AD. Deep learning models have been able to outperform classical methods using auto-encoders and later RotNet-type self-supervised methods \cite{hendrycks2019using,bergman2020deep}. Lately, contrastive learning methods have been used to further improve performance \cite{tack2020csi,reiss2021mean}. A promising line of work suggests to detect anomalies using pretrained  features \cite{reiss2021mean,reiss2021panda}. Pretrained features robustly outperform self-supervised methods, especially on small datasets or when dealing with subtle anomalies. We use similar approaches in this paper.

\begin{table*}[ht!]
  \centering
  \label{tab:phantom_desc}

    \begin{tabular}{lcccccc}
    \toprule      

Classname	&	normal class description 	&	anomaly description		\\
\midrule						
candle	&	burning candle flame	&	flame made to flicker		\\
keyboard	&	sentence being typed	&	different sentence being typed		\\
magnets: knocks	&	graphite oscillating 
between two magnets	&	tabletop is striked		\\
magnets: starting still	&	$\ldots$	&	graphite is still before 
oscillating		\\
magnets: wind	&	$\ldots$	&	wind blows on the graphite		\\
pendulum: motion	&	ball swinging freely	&	swinging is manipulated		\\
pendulum: object	&	$\ldots$	&	ball replaced with other objects		\\
sushi	&	ginger placed atop sushi,
 sushi dipped into sauce	&	 procedure that 
deviates from normal		\\
window	&	window shutters being 
opened at constant speed	&	shutters opened with interruptions		\\
	 \bottomrule
    \end{tabular}
    \caption{PHANTOM Dataset Class Descriptions}
\end{table*}

\textbf{Existing Datasets}
Challenging datasets and benchmarks have been key in advancing the field of computer vision and machine learning. In image anomaly detection, most work has been focused on utilizing standard image classification datasets, while specialized benchmarks focusing on surveillance have been suggested for video anomaly detection \cite{sultani2018real, pranav2020day, rodrigues2020multi, wu2020not}. So far, video AD has so far been heavily reliant on these datasets. On one hand, this ensured that this rapidly-developing field continued to stay relevant to the evaluated task. On the other hand, this may have limited the generality of the methods.

\begin{table}[h]
  \centering
  \label{tab:phantom_eval}

    \begin{tabular}{lcccccc}
    \toprule      

	Classname	&	VIT	&	TimeSformer	&	MNAD	\\
\midrule								
	candle	&	0.76	&	0.71	&	0.49	\\
	keyboard	&	0.94	&	0.95	&	0.89	\\
	magnets: knocks	&	0.52	&	0.41	&	0.18	\\
	magnets: starting still	&	0.79	&	0.64	&	0.42	\\
	magnets: wind	&	0.25	&	0.45	&	0.34	\\
	pendulum: motion	&	0.87	&	1.00	&	0.89	\\
	pendulum: object	&	1.00	&	0.83	&	0.86	\\
	sushi	&	1.00	&	1.00	&	0.61	\\
	window	&	0.86	&	0.90	&	0.24	\\
\midrule								
	Average	&	0.78	&	0.76	&	0.55	\\
	 \bottomrule
    \end{tabular}
    \caption{Anomaly Detection Performance on the PHANTOM dataset (ROCAUC). The results of the image features were obtained using ViT with average pooling and the video features using TimeSformer. These are compared to MNAD surveillance method.}
\end{table}






\section{Evaluated Methods}
\label{sec:methods}
In our work, we use feature extractors from pretrained image and video networks to represent the videos. As the train set contains normal-only videos, we assume that this set makes up a single high-density region in the feature space. With this assumption, for new unseen videos we elect to use the k-nearest neighbors (kNN) distance from the train set as the anomaly score. This simple baseline approach of feature extraction combined with kNN outperforms all previous state-of-the-art methods on small datasets \cite{reiss2021panda}.

\begin{table*}[h]
  \centering
  \label{tab:ss2v_eval}

    \begin{tabular}{lcccccc}
    \toprule      

	Classname	& Subclass	&	ViT w. avg pooling	&	TimeSformer	&	MNAD	\\
\midrule										
	Tearing something into two pieces	&	paper	&	0.70	&	0.51	&	0.36	\\
	Putting something on a surface	&	pen/pencil	&	0.63	&	0.70	&	0.61	\\
	Rolling something on a flat surface	&	pen/pencil	&	0.54	&	0.51	&	0.56	\\
	Something falling like a feather or paper	&	paper	&	0.65	&	0.52	&	0.58	\\
	Unfolding something	&	paper	&	0.64	&	0.51	&	0.44	\\
	Taking one of many similar things on the table	&	pen/pencil	&	0.55	&	0.61	&	0.54	\\
	Closing something	&	box	&	0.42	&	0.45	&	0.58	\\
	Showing that something is empty	&	cup	&	0.51	&	0.46	&	0.48	\\
	Plugging something into something	&	cable	&	0.54	&	0.51	&	0.58	\\
	Pushing something so that it falls off the table	&	pen/pencil	&	0.60	&	0.45	&	0.47	\\
\midrule										
	Average	&		&	0.58	&	0.52	&	0.52	\\
	 \bottomrule
    \end{tabular}
    \caption{Anomaly Detection Performance on the SSv2 dataset (ROCAUC)}
\end{table*}

\subsection{Image features-based AD}

In the first approach, we work with networks pretrained for single-image classification. As these networks have been shown to be very useful on image anomaly detection, the first approach we try is to adapt them to represent entire videos.

The train set $V_{train}=V_1,\ldots,V_n$ is comprised of normal videos only. Each video is separated into frames $V_i=v_{i,1},\ldots,v_{i,t_i}$. 
The frames are evenly sampled and a feature extractor $F$ is used to extract frame features.

\begin{equation}
    F(v_{i,j})=f_{i,j}
\end{equation}

In order to return a single feature vector for each video, we use maximum or average temporal pooling. We refer to each of the above as a time series operation $T$, taking the vector representation of each frame $f_{i,1}, \ldots, f_{i,t_i}$ and returning a single vector representation for the entire video $f_i$.

\begin{equation}
    F(V_i)=T(f_{i,1},\ldots,f_{i,t_i}) = f_i
\end{equation}

We now have a set of embeddings representing the train set, $F_{train}=f_i,\ldots,f_n$. Given a new video sample $V_y$ in the test set, we score its abnormality 
by extracting its features $f_y$ and then by computing its kNN distance from $F_{train}$.

\begin{equation}
    \label{eq:knn}
    d(V_y) = \frac{1}{k} \sum_{f \in N_k(f_y)}{\|f - f_y\|^2}
\end{equation}

Here, $N_k(f_y)$ represents the embeddings of the k-nearest neighbors to $f_y$ in $F_{train}$. After obtaining this distance, we determine whether the video is normal or not by verifying that the distance is greater than some threshold. 
This approach can be viewed as a simplified version of \cite{doshi2020continual,pourreza2021ano}. As official implementations of these methods were not available during this study, we are unable to provide a comparison.

\subsection{Video features-based AD}

Aiming not only to capture the semantics of each image, but also the dynamics of the video, we also utilize pretrained video representations. We use networks pretrained for classification on large video datasets, as such networks are likely to represent a video's content and dynamics in a meaningful way to humans.

Here too, the train set $V_{train}=V_1,\ldots,V_n$ is normal. This time, the feature vector can be extracted directly from the video. 

\begin{equation}
    F(V_i)=f_i
\end{equation}

We use feature extractors to obtain the train set embeddings $F_{train}=f_1,…,f_n$. Given a new test video 
$V_y$, we proceed as discussed above to compute the kNN distance from $F_{train}$ and use it as the anomaly score. 

\subsection{Surveillance method}
To compare our methods to the rapidly developing field of algorithms for AD in surveillance videos, we choose to run the Memory-guided Normality for Anomaly Detection (MNAD \cite{park2020learning}) method on our dataset. MNAD is a state-of-the-art method for video AD for which we were able to find an official implementation. This method returns a score for each video frame. To adapt it such that it returns a single score per video, we take the average of the frame scores.


\section{Experiments}

\label{sec:experiments}
Our experimental results detail when our method performs well and establish a baseline for future work. We compare our method to MNAD on our dataset as well as on Something-Something-V2 \cite{goyal2017something}. Lastly, we run our method on the UCSD Pedestrian 1 and 2 \cite{mahadevan2010anomaly} datasets commonly used in anomaly detection for surveillance.

\subsection{Datasets}

\label{sec:datasets}

\textbf{Physical Anomalous Trajectory or Motion (PHANTOM) Dataset}
To evaluate the presented approaches, we created the Physical Anomalous Trajectory or Motion (PHANTOM) dataset consisting of six classes featuring everyday objects or physical setups, and showing nine different kinds of anomalies. We designed our classes to evaluate detection of various modes of video abnormalities that are generally excluded in video AD settings.  

\begin{table*}[ht!]
  \centering
  \label{tab:transformers_eval}

    \begin{tabular}{lcccccc}
    \toprule      
        	
	classname	&	ResNet w. avg pooling	&	ViT	w. avg pooling &	bLVNet-TAM 
 	&	TimeSformer \\
\midrule										
	candle	&	0.71	&	0.76	&	0.48	&	0.58	\\
	keyboard	&	0.88	&	0.94	&	0.99	&	0.95	\\
	pendulum: motion	&	0.62	&	0.87	&	0.95	&	1.00	\\
	pendulum: object	&	1.00	&	1.00	&	0.92	&	0.87	\\
	sushi	&	1.00	&	1.00	&	1.00	&	1.00	\\
	window	&	0.84	&	0.86	&	1.00	&	0.93	\\
\midrule										
	average	&	0.84	&	0.91	&	0.89	&	0.89	\\
	 \bottomrule
    \end{tabular}
    \caption{Performance of Transformer vs. CNN-based architectures on PHANTOM classes (ROCAUC). 
    }
\end{table*}

The train and test sets of each class contain approximately 30 videos of varying lengths. The train set contains only normal videos, while the test set is evenly balanced between normal and anomalous videos. The classes were designed to be of varying difficulties and to feature different types of anomalies. For example, the \textit{window} class was filmed in multiple lighting scenarios to increase variance.

The normal videos include motion that follows an expected trajectory (\textit{pendulum, keyboard}) or an expected movement (\textit{window}). The \textit{sushi} class features procedural motion, while \textit{candle} and \textit{magnets} feature more subtle motion that only appears locally. The anomalous videos can feature an interference of the regular motion (\textit{window, candle, magnets}), an added or removed step in the usual procedure (\textit{sushi}), motion that follows a different trajectory (\textit{pendulum, keyboard}),  or contains a different object (\textit{pendulum}). The \textit{pendulum} and \textit{magnets} classes contain more than one type of anomaly. An overview of these classes is displayed in Tab.\ref{tab:phantom_desc}.

\textbf{General Activity Dataset}
\label{sec:ssv2}
To examine a harder, even more general video AD setting, we work with the Something-Something-V2 (SSv2) dataset. It features 174 classes of various activities such as \textit{Throwing something in the air and catching it} where \textit{something} is not limited to a single object type. We adapt this dataset for use for general AD as described in Sec. \ref{sec:exp_general} and it can be thought of as suitable crossover between physical and surveillance videos. 

\begin{figure}[t]
\begin{center}
\begin{tabular}{@{\hskip0pt}c@{\hskip5pt}c@{\hskip5pt}c@{\hskip5pt}c@{\hskip5pt}c@{\hskip5pt}c@{\hskip0pt}c@{\hskip0pt}c@{\hskip0pt}c}

Tearing & Closing\\

\includegraphics[width=0.45\linewidth]{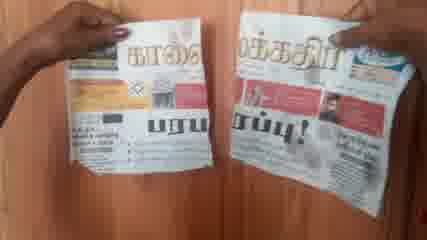} &
\includegraphics[width=0.45\linewidth]{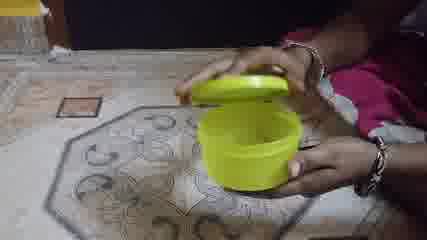} \\
\includegraphics[width=0.45\linewidth]{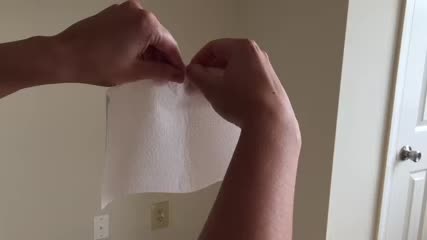} &
\includegraphics[width=0.45\linewidth]{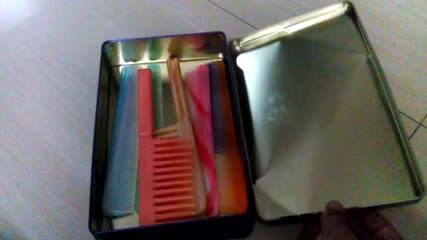} \\

\end{tabular}
\end{center}
\caption{Examples of frames from two of the ten video types in the SSv2 classes that we used. For each class, two frames are shown to demonstrate the high class variability.}
\label{fig:ffhq}
\vspace{-1em}
\end{figure}

\subsection{Physical AD on PHANTOM dataset}

To determine the ability of our proposed method at finding anomalies in videos, we test it on our dataset. We explore the different modes of the method and uncover the strengths and weaknesses of each setting. We measure our results using the area under the curve of the receiver operating characteristic (ROCAUC) as is common in previous works \cite{reiss2021panda,hendrycks2019using}.

\textbf{Image Features: }Here we run the videos in our dataset frame-wise through different pretrained image networks. We choose a Residual Network (ResNet \cite{he2016deep}) and the Vision Transformer (ViT \cite{dosovitskiy2020image}) as previously they have been tested extensively on numerous image datasets and have been shown to achieve excellent transfer learning results. Each network comes pretrained on ImageNet and ImageNet-21K. In order to summarize the frame features into a single feature vector we use temporal pooling. After obtaining the video features, we give each video an anomaly score. The results of the best performing image-based method on the PHANTOM dataset appear in Tab.\ref{tab:phantom_eval}. 

\textbf{Video Features: }Video networks return a single feature vector per video, therefore it is not necessary to pool the frame-level features. We use the following pretrained video networks: TimeSformer \cite{bertasius2021space} pretrained on Kinetics-600 \cite{carreira2018short} and SSv2, and the Big-Little-Video-Net Temporal Aggregation Module (bLVNet-TAM \cite{fan2019more}) architecture pretrained on SSv2. A summary of the best-performing method is given in Tab.\ref{tab:phantom_eval}. 

\textbf{Surveillance Method: } Lastly, we must compare the performance of the image and video features to a method meant for surveillance footage. For this, We choose MNAD and train it on the normal classes for 30, 60 and 120 epochs. We obtain the best performing run with 60 epochs. The results are shown in Tab.\ref{tab:phantom_eval}.

\begin{table}[t!]
  \centering

    \begin{tabular}{lcccccc}
    \toprule      
	Method	&	UCSD Ped1	&	UCSD Ped2	\\
\midrule						

	Kim\cite{kim2009observe}	&	59.0	&	69.3	\\
	Liu\cite{liu2018future}	&	83.1	&	95.4	\\

	Ionescu\cite{ionescu2019object}	&	-	&	97.8	\\
			Pang\cite{pang2020self}	&	71.7	&	83.2	\\
		MNAD\cite{park2020learning}	&	-	&	97.0	\\
	Ours	&	52.6	&	75.0	\\
	 \bottomrule
    \end{tabular}
    \caption{Comparison to other methods on UCSD Ped1 and Ped2 datasets}
      \label{tab:ucsd_eval}

\end{table}

\subsection{Comparison to MNAD on SSv2}
\label{sec:exp_general}

Next, we choose to contrast our simple method of feature extraction and kNN to MNAD on the SSv2 dataset. We choose 10 classes and from among these classes, we determine the largest subclass and label it normal. We run this subclass against the videos in the largest subclass of the other 9 classes which are labeled anomalous. This evaluation is similar to that of the PHANTOM dataset. The names of the chosen classes and subclasses appear in Tab.\ref{tab:ss2v_eval} together with the results.
This task proved challenging for all methods and the results show that while our method works well on our PHANTOM dataset, it is not well-adapted for general videos. This highlights general and physical videos as different modalities which may call for different AD solutions.

\subsection{Comparison on surveillance dataset}

In order to examine the performance of our proposed benchmarks against surveillance videos, we test it on other types of data. To this end, we select the UCSD Pedestrian 1 and 2 datasets. These datasets contain surveillance footage of two pedestrian scenes and are commonly used for video AD evaluation.

Surveillance AD methods differ from ours in that each frame is given an anomaly score. To overcome this, 
we adapt our procedure in the following way. We break the video into overlapping windows of constant length such that each frame appears in several windows. Thus, each video in the train set contributes several feature vectors equal to the number of overlapping windows in that video. Given an unseen video, we again divide the video into windows and compute the window features. We compute the kNN distance between each window and the windows in the train set. As each frame can appear in several windows, the anomaly score that is given is the average of the distances computed for the windows in which that frame appears. A summary of the results appears in Tab.\ref{tab:ucsd_eval} and are discussed in Sec. \ref{sec:discussion}. We find that the better performing method on physical and general benchmarks under-performs on surveillance videos. This emphasizes the different nature of these tasks.






\section{Discussion}
\label{sec:discussion}

\textbf{Video features vs. image features:} Video features often outperformed the image features with the exception of the features obtained by ViT pretrained on ImageNet-1K. Still, we note that among the classes that favored the image features were those that either had subtle differences in the motion of the normal and anomalous classes (\textit{candle, magnets}) or had similar motion but feature an entirely different object (\textit{pendulum}).  Conversely, the classes that favored the video features were those that feature an interruption in the motion (\textit{window}) or motion that follows a different trajectory (\textit{keyboard, sushi}). Intuitively, this is in line with what one may expect as the image networks are pretrained to classify objects of different classes while the video networks are pretrained to classify different types of motion. This may serve as a general guideline in order to determine when one technique may outperform the other. 



\textbf{The need for a general video AD method:} While our method works well at finding dynamics-related anomalies like those seen in our dataset, the results in Tab.\ref{tab:ucsd_eval} show that our pretrained methods are not well-adapted to surveillance datasets. On the other hand, while MNAD attains state-of-the-art results on UCSD, it is not able to solve our dataset as easily. When running these methods on the SSv2 dataset,  Tab.\ref{tab:ss2v_eval} shows that they both struggle with this task. The videos in the train and test sets that we chose contain high variability both in the video background and the object in question. For example, the \textit{Tearing paper} subclass is filmed in numerous scenes with the paper being either blank or with words printed on it, and being of different sizes and colors. This variability made solving these classes difficult for both our method and MNAD. These results stress the difference between these video AD settings and the need to develop a technique that will be able to solve general video AD.

\section{Conclusion}
In our work, we focus on detecting anomalies in physical and general videos. We introduce new such benchmarks and evaluate baseline approaches on them. We find that the simple pretrained approach that struggles to outperform on surveillance data outperforms on our own suggested benchmarks. Taken together, our work highlights physical and general video anomaly detection as new tasks that call for the development of new approaches.

\section{Acknowledgements}

This work was partly supported by the Federmann Cyber Security Research Center in conjunction with the Israel National Cyber Directorate.

\bibliographystyle{IEEEbib}
\bibliography{refs}

\begin{thebibliography}{10}

\bibitem{reiss2021panda}
Tal Reiss, Niv Cohen, Liron Bergman, and Yedid Hoshen,
\newblock ``Panda: Adapting pretrained features for anomaly detection and
  segmentation,''
\newblock in {\em Proceedings of the IEEE/CVF Conference on Computer Vision and
  Pattern Recognition}, 2021.

\bibitem{hendrycks2019using}
Dan Hendrycks, Kimin Lee, and Mantas Mazeika,
\newblock ``Using pre-training can improve model robustness and uncertainty,''
\newblock in {\em International Conference on Machine Learning}. PMLR, 2019.

\bibitem{tack2020csi}
Jihoon Tack, Sangwoo Mo, Jongheon Jeong, and Jinwoo Shin,
\newblock ``Csi: Novelty detection via contrastive learning on distributionally
  shifted instances,''
\newblock {\em arXiv:2007.08176}, 2020.

\bibitem{luo2017revisit}
Weixin Luo, Wen Liu, and Shenghua Gao,
\newblock ``A revisit of sparse coding based anomaly detection in stacked rnn
  framework,''
\newblock in {\em Proceedings of the IEEE International Conference on Computer
  Vision}, 2017.

\bibitem{mahadevan2010anomaly}
Vijay Mahadevan, Weixin Li, Viral Bhalodia, and Nuno Vasconcelos,
\newblock ``Anomaly detection in crowded scenes,''
\newblock in {\em 2010 IEEE Computer Society Conference on Computer Vision and
  Pattern Recognition}. IEEE, 2010.

\bibitem{georgescu2021anomaly}
Mariana-Iuliana Georgescu, Antonio Barbalau, Radu~Tudor Ionescu, Fahad~Shahbaz
  Khan, Marius Popescu, and Mubarak Shah,
\newblock ``Anomaly detection in video via self-supervised and multi-task
  learning,''
\newblock in {\em Proceedings of the IEEE/CVF Conference on Computer Vision and
  Pattern Recognition}, 2021.

\bibitem{park2020learning}
Hyunjong Park, Jongyoun Noh, and Bumsub Ham,
\newblock ``Learning memory-guided normality for anomaly detection,''
\newblock in {\em Proceedings of the IEEE/CVF Conference on Computer Vision and
  Pattern Recognition}, 2020.

\bibitem{ionescu2019object}
Radu~Tudor Ionescu, Fahad~Shahbaz Khan, Mariana-Iuliana Georgescu, and Ling
  Shao,
\newblock ``Object-centric auto-encoders and dummy anomalies for abnormal event
  detection in video,''
\newblock in {\em Proceedings of the IEEE/CVF Conference on Computer Vision and
  Pattern Recognition}, 2019.

\bibitem{gong2019memorizing}
Dong Gong, Lingqiao Liu, Vuong Le, Budhaditya Saha, Moussa~Reda Mansour, Svetha
  Venkatesh, and Anton van~den Hengel,
\newblock ``Memorizing normality to detect anomaly: Memory-augmented deep
  autoencoder for unsupervised anomaly detection,''
\newblock in {\em Proceedings of the IEEE/CVF International Conference on
  Computer Vision}, 2019.

\bibitem{hasan2016learning}
Mahmudul Hasan, Jonghyun Choi, Jan Neumann, Amit~K Roy-Chowdhury, and Larry~S
  Davis,
\newblock ``Learning temporal regularity in video sequences,''
\newblock in {\em Proceedings of the IEEE conference on computer vision and
  pattern recognition}, 2016.

\bibitem{lee2019bman}
Sangmin Lee, Hak~Gu Kim, and Yong~Man Ro,
\newblock ``Bman: bidirectional multi-scale aggregation networks for abnormal
  event detection,''
\newblock {\em IEEE Transactions on Image Processing}, 2019.

\bibitem{pang2020self}
Guansong Pang, Cheng Yan, Chunhua Shen, Anton van~den Hengel, and Xiao Bai,
\newblock ``Self-trained deep ordinal regression for end-to-end video anomaly
  detection,''
\newblock in {\em Proceedings of the IEEE/CVF Conference on Computer Vision and
  Pattern Recognition}, 2020.

\bibitem{bergman2020deep}
Liron Bergman, Niv Cohen, and Yedid Hoshen,
\newblock ``Deep nearest neighbor anomaly detection,''
\newblock {\em arXiv:2002.10445}, 2020.

\bibitem{reiss2021mean}
Tal Reiss and Yedid Hoshen,
\newblock ``Mean-shifted contrastive loss for anomaly detection,''
\newblock {\em arXiv:2106.03844}, 2021.

\bibitem{sultani2018real}
Waqas Sultani, Chen Chen, and Mubarak Shah,
\newblock ``Real-world anomaly detection in surveillance videos,''
\newblock in {\em Proceedings of the IEEE conference on computer vision and
  pattern recognition}, 2018.

\bibitem{pranav2020day}
Mantini Pranav, Li~Zhenggang, et~al.,
\newblock ``A day on campus-an anomaly detection dataset for events in a single
  camera,''
\newblock in {\em Proceedings of the Asian Conference on Computer Vision},
  2020.

\bibitem{rodrigues2020multi}
Royston Rodrigues, Neha Bhargava, Rajbabu Velmurugan, and Subhasis Chaudhuri,
\newblock ``Multi-timescale trajectory prediction for abnormal human activity
  detection,''
\newblock in {\em Proceedings of the IEEE/CVF Winter Conference on Applications
  of Computer Vision}, 2020.

\bibitem{wu2020not}
Peng Wu, Jing Liu, Yujia Shi, Yujia Sun, Fangtao Shao, Zhaoyang Wu, and Zhiwei
  Yang,
\newblock ``Not only look, but also listen: Learning multimodal violence
  detection under weak supervision,''
\newblock in {\em European Conference on Computer Vision}. Springer, 2020.

\bibitem{doshi2020continual}
Keval Doshi and Yasin Yilmaz,
\newblock ``Continual learning for anomaly detection in surveillance videos,''
\newblock in {\em Proceedings of the IEEE/CVF Conference on Computer Vision and
  Pattern Recognition Workshops}, 2020.

\bibitem{pourreza2021ano}
Masoud Pourreza, Mohammadreza Salehi, and Mohammad Sabokrou,
\newblock ``Ano-graph: Learning normal scene contextual graphs to detect video
  anomalies,''
\newblock {\em arXiv:2103.10502}, 2021.

\bibitem{goyal2017something}
Raghav Goyal, Samira Ebrahimi~Kahou, Vincent Michalski, Joanna Materzynska,
  Susanne Westphal, Heuna Kim, Valentin Haenel, Ingo Fruend, Peter Yianilos,
  Moritz Mueller-Freitag, et~al.,
\newblock ``The" something something" video database for learning and
  evaluating visual common sense,''
\newblock in {\em Proceedings of the IEEE international conference on computer
  vision}, 2017.

\bibitem{he2016deep}
Kaiming He, Xiangyu Zhang, Shaoqing Ren, and Jian Sun,
\newblock ``Deep residual learning for image recognition,''
\newblock in {\em Proceedings of the IEEE conference on computer vision and
  pattern recognition}, 2016.

\bibitem{dosovitskiy2020image}
Alexey Dosovitskiy, Lucas Beyer, Alexander Kolesnikov, Dirk Weissenborn,
  Xiaohua Zhai, Thomas Unterthiner, Mostafa Dehghani, Matthias Minderer, Georg
  Heigold, Sylvain Gelly, et~al.,
\newblock ``An image is worth 16x16 words: Transformers for image recognition
  at scale,''
\newblock {\em arXiv:2010.11929}, 2020.

\bibitem{bertasius2021space}
Gedas Bertasius, Heng Wang, and Lorenzo Torresani,
\newblock ``Is space-time attention all you need for video understanding?,''
\newblock {\em arXiv:2102.05095}, 2021.

\bibitem{carreira2018short}
Joao Carreira, Eric Noland, Andras Banki-Horvath, Chloe Hillier, and Andrew
  Zisserman,
\newblock ``A short note about kinetics-600,''
\newblock {\em arXiv:1808.01340}, 2018.

\bibitem{fan2019more}
Quanfu Fan, Chun-Fu Chen, Hilde Kuehne, Marco Pistoia, and David Cox,
\newblock ``More is less: Learning efficient video representations by
  big-little network and depthwise temporal aggregation,''
\newblock {\em arXiv:1912.00869}, 2019.

\bibitem{kim2009observe}
Jaechul Kim and Kristen Grauman,
\newblock ``Observe locally, infer globally: a space-time mrf for detecting
  abnormal activities with incremental updates,''
\newblock in {\em 2009 IEEE conference on computer vision and pattern
  recognition}. IEEE, 2009.

\bibitem{liu2018future}
Wen Liu, Weixin Luo, Dongze Lian, and Shenghua Gao,
\newblock ``Future frame prediction for anomaly detection--a new baseline,''
\newblock in {\em Proceedings of the IEEE conference on computer vision and
  pattern recognition}, 2018.

\end{thebibliography}

\end{document}